\documentclass[acmconf, nonacm]{acmart}
\AtBeginDocument{%
  \providecommand\BibTeX{{%
    \normalfont B\kern-0.5em{\scshape i\kern-0.25em b}\kern-0.8em\TeX}}}

\begin{document}

\title{ReGNL: Rapid Prediction of GDP during Disruptive Events using Nightlights}


\author{Rushabh Musthyala}
\authornote{The authors have made equal contribution to the research}
\affiliation{%
  \institution{BITS Pilani, Hyderabad Campus}
  \country{India}
}
\email{f20180433@hyderabad.bits-pilani.ac.in}
\author{Rudrajit Kargupta}
\authornotemark[1]
\affiliation{%
  \institution{BITS Pilani, Hyderabad Campus}
  \country{India}
}
\email{f20170452@hyderabad.bits-pilani.ac.in}
\author{Hritish Jain}
\authornotemark[1]
\affiliation{%
  \institution{BITS Pilani, Hyderabad Campus}
  \country{India}
}
\email{f20190356@hyderabad.bits-pilani.ac.in}
\author{Dipanjan Chakraborty}
\affiliation{%
  \institution{BITS Pilani, Hyderabad Campus}
  \country{India}
}
\email{dipanjan@hyderabad.bits-pilani.ac.in}

\newcommand{\rushabh}[1]{\textcolor{blue}{{\emph{ (Rushabh:)  #1}}}}

\renewcommand{\shortauthors}{Musthyala, et al.}

\begin{abstract}

Policy makers often make decisions based on parameters such as GDP, unemployment rate, industrial output, etc. The primary methods to obtain or even estimate such information are resource intensive and time consuming. In order to make timely and well-informed decisions, it is imperative to be able to come up with proxies for these parameters which can be sampled quickly and efficiently, especially during disruptive events, like the COVID-19 pandemic.
Recently, there has been a lot of focus on using remote sensing data for this purpose. The data has become cheaper to collect compared to surveys, and can be available in real time. In this work, we present Regional GDP NightLight (ReGNL), a neural network based model which is trained on a custom dataset of historical nightlights and GDP data along with the geographical coordinates of a place, and estimates the GDP of the place, given the other parameters. Taking the case of 50 US states, we find that ReGNL is disruption-agnostic and is able to predict the GDP for both normal years (2019) and for years with a disruptive event (2020). ReGNL outperforms timeseries ARIMA methods for prediction, even during the pandemic. Following from our findings, we make a case for building infrastructures to collect and make available granular data, especially in resource-poor geographies, so that these can be leveraged for policy making during disruptive events.
\end{abstract}




\maketitle

\section{Introduction}
\label{1}

During disruptive events, like the COVID-19 pandemic, it becomes important for policy makers to quickly estimate the impact of the event on various parameters like GDP (Gross Domestic Product), unemployment rate, industrial output, etc. Existing methods to collect or estimate these parameters are time and labour intensive as they require extensive surveys and data collection. Therefore, computation and publication of these data takes time, and policy making is often done in hindsight. While the existing models are time tested and work well in normal times, they fall short during disruptive events like the COVID-19 pandemic, when policy makers need methods with quick turnaround time to estimate economic parameters and subsequently make policy decisions.


In this paper we present \textit{ReGNL} (Regional GDP-NightLight), a neural networks based technique to rapidly predict the GDP of a given geography using remote sensing. The remote sensing data we use is \textit{Nightlights}, processed from data collected by sensors on board earth-orbiting satellites. Nightlights refer to the visible lights being emitted by activities on earth's surface, remotely sensed by sensors on-board various satellites. With advances in data storage and data processing techniques, and Machine Learning algorithms, it is possible to leverage the large amounts of remote sensing datasets for estimation of macroeconomic parameters like GDP.

Off-late, nightlight has emerged as a convenient proxy used to ascertain economic activity of a region \cite{10.1257/jep.30.4.171, 10.1145/3394486.3403347}. We develop ReGNL to estimate the GDP of a given geography using the nightlights data from the same geography. We present our technique on datasets from different states in the USA, and also demonstrate its applicability on the provinces of Germany, because of the availability of regular data from these countries. However, our techniques are general and can be extended to other geographies (cities, states and countries) as well, given that granular and regular data is available. Our model (ReGNL) is trained on the quarterly nightlights and GDP data from 50 US states from 2014 to 2018. We test ReGNL for a normal (economy wise) timeframe by predicting the GDP for the year 2019 (weighted error: 0.7066, weighted error is explained in Section \ref{4.2}). Next, we test ReGNL for resilience against disruptive events by considering Quarter 2 (April to June) of 2020 as the disruption timeframe. ReGNL is able to predict the GDP for the 50 US states for Quarter 2 of 2020 with a weighted error of 0.7243 from the actual GDP values. In comparison, when we use a time series analysis using ARIMA as the baseline prediction model, we obtain a weighted error of 1.9488 for the same quarter. ReGNL therefore outperforms the timeseries ARIMA model for a timeframe with a disruptive event.

The novelty of our technique is in the curation of a custom dataset comprising of around 1000 datapoints which combines the nightlight values, GDP values, along with the geographical coordinates in the form of latitude and longitude (centroid) of the states. The inclusion of geographical coordinates in the dataset is able to introduce an aspect of location proximity to our model which is not captured by the nightlights data alone.


The salient contributions of our work are:
\begin{enumerate}
    \item We curate a custom dataset combining the quarterly nightlights values, quarterly GDP values and latitude and longitude of 50 US states from 2014-2020. We also curate a second dataset with nightlights and GDP values from Germany, India, China and Spain.
    \item We develop the ReGNL (Regional GDP-NightLight Forecasting) model, a neural network which predicts the GDP of a geography using the nightlights and latitude and longitude as features. ReGNL is able to estimate the GDP with a consistent weighted error of just 0.6981, on an average.
    \item Proof that with the enough data, nightlights can be used to make estimations agnostic of disruptive events.
\end{enumerate}
Our results validate previous claims that nightlight can be used as a proxy for GDP, and further asserts that it is robust even during disruptive events, when conventional data collection methods may not be possible.


The rest of the paper is structured as follows - Section \ref{2} contexualizes our work within existing research in the domain. Section \ref{3} gives a detailed description of the dataset curation process. Section \ref{4} provides a detailed description of the approach and model we develop as well as the details of the results obtained. Finally conclusions are drawn and future research directions are given in section \ref{5}.

\section{Related Work}
\label{2}

Several researchers have explored using remote sensing data in various applications in Economics. Bansal et al. \cite{compass-1, compass-2} curate custom datasets from India which can be used for various policy decisions. Robinson et al. build a model to understand changes in building patterns longitudinally \cite{compass-3}. Donaldson, et al. \cite{10.1257/jep.30.4.171} summarise some existing work in the domain of economics using satellite imagery. They talk about the scope that exists for further research in the field of Economics using remote sensing. The paper also discusses the various shortcomings of using remote sensing for research in the economic context, such as spatial dependence, privacy concerns, dataset size and measurement error. This helps researchers understand the progress that has been made so far in this domain and about the gap that exists. Xie, et al. \cite{10.5555/3016387.3016457} highlight the problems faced by policymakers due to lack of reliable data to base their decisions on. The paper suggests the use of satellite image data to gauge the socio-economic status of different regions. However, training Deep Learning models on image data requires a huge amount of data and the training data available is very limited. Hence, the authors use a Transfer Learning framework which involves using a model trained for a separate task/dataset to be reused in a different context. In this case the nightlight data is used as a proxy. The accuracy of the model was found to be close to that of actual field surveys. Liu, et al. \cite{liu2021nightlight} also use a transfer learning framework on the premise that nightlights are an effective proxy for the economic activity of a region (Mainland China in this case). The authors use satellite images along with nightlight intensity levels as the input set of features to train their model. Eventually, they were able to predict the 2018 GDP with a reasonably high R-squared value of 0.71. Our work is different in the sense that (i) we do not use a transfer learning framework, and, (ii) we do not use satellite ``image'' data, instead we collect and compile remote sensing nightlight data for a region and train our model (ReGNL) on the same. Our motivation for doing this is because it is computationally cheaper to train a model on the nightlight (numeric) values instead of training it on images. Furthermore, Computer Vision tasks become susceptible to issues regarding image quality, saturation, blurring, etc. 

Han, et al. \cite{10.1145/3394486.3403347} in their work present a three step approach for judging the economic status of a region without having access to ground truth figures by using high resolution satellite images: (i) the man-made and natural objects in a satellite image are classified (segregated) into different collections by using a clustering framework, (ii) partial order graph of the collections identified in the previous step are developed, and, finally, (iii) a Convolutional Neural Network (CNN) based framework is used to sort each of the satellite images (grids) on the basis of the relative positioning of the collections. Eventually a score is assigned to each of the locations (satellite images), this score is indicative of the near real-time economic development of that corresponding location. The robustness of the methodology proposed in this paper is demonstrated by applying the same methodology to different economies such as Vietnam (a developing economy), South Korea (a developed economy) and Malawi. Similar to the previous paper, this paper is also based on satellite ``image'' data, whereas our work here is based on processed (numeric) nighlight data, in order to avoid the computational complexity associated with processing satellite images. Otchia, et al. \cite{otchia2020industrial} use Machine Learning methodologies on nighttime data to study the industrial progress in Africa. This research further confirms the validity of using nighttime data for gauging the economic progress for regions where there is a gap in the data or the quality of data is poor.

Asher, et al. \cite{10.1093/wber/lhab003} found a statistically significant relationship between nightlight data and economic variables such as, per-capita income and consumption, electricity coverage, population growth and density. The analysis done in this paper is based on about 6 lakh locations across India. This validates our assumption of using nightlight data for determining, or using as a proxy for, the economic parameters of a region. Finally, the authors identify the poverty distribution across India and point out that poverty alleviation programmes will have more impact if implemented at the (much more granular) village level as opposed to the district level. Dasgupta, et al. \cite{PPR:PPR260387} use Nightlight Data in combination with electricity consumption data to predict the impact of COVID-19 on the Indian economy. This analysis was done at a national level and predicted a contraction of 24 percent in in the 1st Quarter of the Financial Year 2020, which was quite close to the official figure of -23.9 percent. However, the model performed below par on the state level. This is one of the gaps that our work aims to fill. Our work focuses on much more specific, state or provincial levels so as to help policy-making decisions with the predictions being reported at a much higher resolution (granularity) about each state or province, rather than it being reported just at the national level. Ustaoglu, et al. \cite{ustaoglu2020spatial} develop a pixel-level agricultural and non-agricultural GDP map for Turkey. The authors use various datasets, combining the Terra MODIS-Enhanced Vegetation Index, the VIIRS-NPP nighttime image data and land use/cover data from CORINE. Including the vegetation information and land cover data leads to a marked improvement in the GDP estimates since agriculture is a major driving force in Turkey's economic output. The work done by the authors in this paper outline the significant impact the remotely sensed information can make in GDP estimation and any other economic indicator. 
Kulkarni et al. \cite{kulkarni2011revisiting} use satellite image data, to obtain nightlight information, to make estimates of the economic activity at the sub-regional levels for the nations: India, USA and China. These estimates were then compared with the sub-regional economic indicators. This analysis was done for China, USA and India for the years 2001 to 2007 and the results were quite encouraging for USA and India but not for China. The unfavourable results obtained for China, were attributed to various factors, such as, the saturation level of satellite sensors, the internal migration in China, etc. Skoufias et al. \cite{skoufias2021can} use VIIRS data (which is a format of nightlight data) to gauge the impact that natural disasters such as floods, earthquakes etc. have on the nightlight intensity of a region and how do these correspond to the short-term damage caused there. The authors conclude by apprising researchers to be careful while using the VIIRS data to critique the consequences of natural hazards.

Gallup, et al. \cite{gallup1999geography} examined the role that the geography of a region plays in its economic progress. In their work they highlight how the location and climate of a region impacts its economy through different facets such as agricultural productivity, transportation costs and the burden of disease. In several locations where the economic growth has been slow, the problem has been further compounded by an increased rate of growth in population. Yet another unfavourable situation for countries are when they are landlocked and located in a region with lack of access to coasts, this leads to a reduced amount of international trade which is one of the important drivers for economic growth. These disadvantaged regions are the ones that are predicted to witness the majority of the global population growth. The hypotheses proposed in this paper acts as the basis for us to include geographical coordinates using longitude and latitude as one of the input features in our analysis so as to enhance the working of our model, ReGNL. Maskell, et al. \cite{doi:10.1080/00291959608542835} talk about majority of a region's economic activity being driven by organisations embedded in that region's structure and that their economic growth is reliant on competencies specific to them, despite the increased levels of globalization, worldwide. Bickenbach, et al. \cite{bickenbach2016night} in their work highlight that the relationship between the growth in nightlight and the growth in GDP varies significantly, economically as well as statistically, across regions. This was established using regional data from Brazil and India. Therefore, trying to draw a relationship between the two without accounting for the positional information (such as longitude and latitude) may lead to poor results. These papers outline the importance of including geographical location in our analysis.

Another possible use of this analysis can be to verify GDP estimates released by various government bodies which have been known to inflate their findings. Martinez, Luis R. \cite{martinez2018much} in his work finds that most authoritarian regimes inflate their growth rate findings by a factor of about 1.15 to 1.3. It was found that the misrepresentation of GDP statistics in authoritarian or autocratic regimes is higher when there is a higher incentive to exaggerate economic well being of a region. Some of these could be just before elections or to hide a sustained dip in economic growth. Using our work and other similar methodologies these GDP findings can be scrutinised with a reasonable level of accuracy.

All of the work cited in this section points towards the feasibility of using remotely sensed nightlight data to reasonably estimate the economic output of a region. We hence believe that nightlights can be used for overcoming the lack of real-time data for assessing the economic health of a region and proceed towards extending this idea to disruptive events.

\section{Dataset}
\label{3}

In this section we describe the data collection and processing pipeline. There are 3 components to the dataset: nightlight values, GDP values and geographical coordinates in the form of latitudes and longitudes.

\subsection{Nightlights Data and Geographic Coordinates}\label{sec:nightlights}
Several remote sensing datasets and tools such as the SHRUG open data tool \cite{almn2021}, USGS Earth Explorer Tool \cite{usgs} and the Google Earth Engine \cite{earthengine} were explored. The Google Earth Engine tool was chosen as it had some specific advantages over the others: Google Earth Engine has a rich repository of datasets collected from different verified sources; the datasets are preprocessed and ready for use in different research contexts; it comes with an in-built code editor and library which enables researchers to write code in JavaScript and easily use the predefined functions in the library.


The Google Earth Engine, as outlined in the previous paragraph, is just a tool and not the dataset itself. The dataset that we use in this work for obtaining the remote-sensing data is provided by ``Earth Observation Group, Payne Institute for Public Policy, Colorado School of Mines'' \cite{doi:10.1080/01431161.2017.1342050}. The dataset contains nightlight data in the form of Average Day Night Band (DNB) radiance. The average DNB radiance band, part of the VIIRS dataset, is a metric that ranges from -1.5 to 193565 units and is reported in $\frac{nanoWatts}{cm^{2}/sr}$. This data is from the Visible and Infrared Imaging Suite (VIIRS) Day Night Band (DNB) on board the JPSS (Joint Polar Satellite System) satellites. The VIIRS-DNB has shown a significant improvement in capturing low light images as compared to the Defense Meteorological Satellite Program (DMSP) satellites which were mounted with the Operational Linescan Sensors (OLS). Elvidge, et al. \cite{articleVIIRS} conclude in their work the superiority of VIIRS data as compared to DMSP data based on these parameters: dynamic range, spatial resolution, calibrations, quantization and the
availability of spectral bands suitable for discrimination of thermal sources of light emissions. Similarly Gibson, et al. \cite{GIBSON2021102602} show the superiority of VIIRS data over DMSP data by comparing the two data sources (VIIRS and DMSP) by using them for GDP predictions. The predictive performance of DMSP data was found to be significantly lower than the VIIRS data particularly for rural areas which have lower population densities and at lower levels of spatial hierarchy like at the country level. The relationship between GDP and city lights was found to be twice as noisy in the DMSP data than the VIIRS data. The VIIRS data was found to be superior to the DMSP data in several aspects including lack of proper calibration, top-coding and blurring. This furthers our reasoning for using the newer and better VIIRS dataset over the DMSP dataset.

Shapefiles contain the geometric information about the boundaries of a region like country or state. When overlaid on a map, shapefiles can be used to identify the states or provinces we want to study. The shapefiles for our analysis were obtained from DIVA-GIS \cite{diva-gis} and the US National Weather Service \cite{weatherShapeFile}. Using the VIIRS dataset and the shapefiles we obtain the nightlight information mapped to each of corresponding state along with the state's geographic coordinates in the form of latitudes and longitudes.

\begin{figure*}[hbt!]
  \centering
  \includegraphics[width=\textwidth]{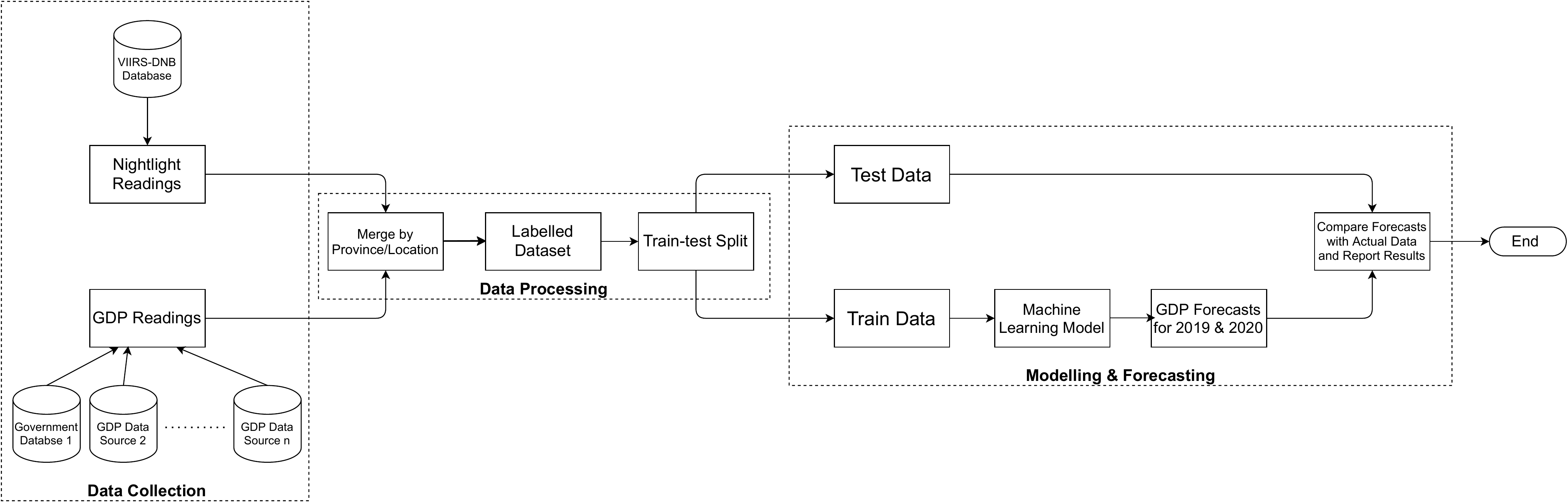}
  \caption{The proposed methodology.}
  \label{fig: 2}
\end{figure*}

\subsection{GDP Data}
The dataset obtained Section \ref{sec:nightlights} has the State or Province Name, Mean Nightlight Value, Latitude, Longitude with one observation for every month. These features need to be mapped with the GDP values for each of the corresponding provinces or states. The quarterly GDP values for the United States of America were obtained from Bureau of Economic Analysis \cite{bea}. Subsequently, we averaged the monthly nightlight values into quarters and mapped them to their corresponding GDP values. Furthermore, GDP Data is reported with respect to a particular base year. This to allow for purchasing power comparisons and to be able to report growth of a region while adjusting for inflation. These base years are updated from time to time which is an issue for our analysis because we need the entire dataset to have a common base year so as to have a uniformity in the dataset, and to ensure that the economic activity for a given region is measured on a common scale for all the regions in our dataset. To address this issue, the base year was chosen to be 2011 and adjusted the GDP values of all the regions accordingly.



\begin{table*}
\caption{Structure of our dataset}
\label{table: 6}
\begin{tabular}{|l|l|l|l|l|l|l|}
\hline
State/Province & Year & Quarter & Latitude & Longitude & Mean Nightlight & GDP   \\
\hline
California           & 2014 & Q1      & 37.2453   & -119.6081    & 1.4385           & 2252602.814 \\
Alabama           & 2014 & Q1      & 32.7935   & -.826786    & 1.4918           & 183844.57 \\
Virginia          & 2018 & Q4      & 37.5164   & -78.829 6   & 2.0349           & 475154.764 \\ 
\hline
\end{tabular}
\end{table*}


A snapshot of the final dataset is in Table \ref{table: 6}. While the main focus of our study was the United States, we were also able to collect annual GDP data for Spain \cite{countryeconomy}, Germany \cite{Statisticsportal}, India \cite{mspi} and China \cite{nbs-china}. Since these countries follow different cycles for their financial year GDP reporting, we averaged out the nightlight data as per their respective financial years. We performed similar experiments on the data from Germany, Spain, India and China. Out of these countries, at the time of our analysis, the annual GDP data for 2020 was only available for Germany, to verify our findings. We report the results of applying our model on Germany in Section \ref{4.4}. We show the pipeline for dataset creation and modelling in Figure \ref{fig: 2}.

\section{Experiments, Model and Results}
\label{4}


We aim to answer the following with regards to ReGNL:
\begin{enumerate}
    \item Does it work for both regular (economy-wise) years and years with disruptive events?
    \item Does the inclusion of geographical coordinates in the form of latitude and longitude result in improvement in its performance? 
\end{enumerate}

\subsection{Timeframe with Disruptive Event}
\begin{figure}[hbt!]
  \centering
  \includegraphics[width=\linewidth]{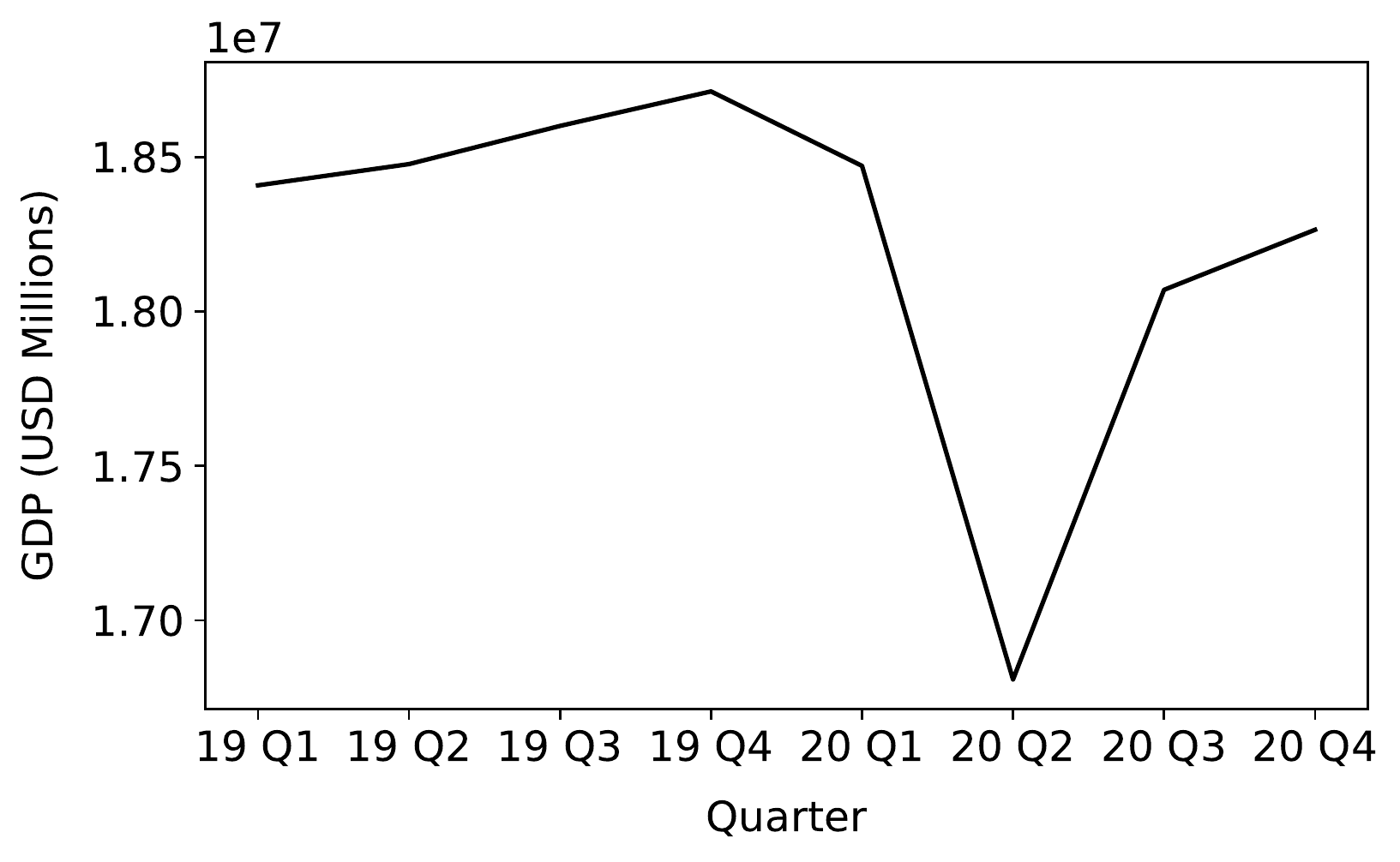}
  \caption{USA GDP 2019-20 (indicating the economic dip in 2020 Q2)}
  \label{fig: 1}
\end{figure}

The US GDP for the years 2019 and 2020 is plotted in Figure \ref{fig: 1}. The United States experienced a significant drop in the national GDP during the 2nd quarter of 2020, which can be attributed to the COVID - 19 pandemic. To test our method across regular timeframes and timeframes with disruptive events, we trained ReGNL on data from 2014-2018 and used it to make predictions for 2019 (regular year) and 2020 (year with a disruptive event). 

\subsection{Evaluating the Incorporation of Geographic Coordinates}
\label{4.2}
To evaluate the value added by incorporating geographical coordinates in the dataset, ReGNL was trained under 2 different scenarios - one containing just the mean nightlight of a region and the other consisting of the mean nightlight as well as the latitude and longitude of the region. The metric used to compare different predictions was a weighted error, in which the error of each state was weighted by its contribution to the total GDP of the country. The reason behind using a weighted error was to ensure that adequate importance is given to states in accordance with how high or low their GDP contribution is to the country. The error is computed using Equation \ref{eqn:1}.

\begin{equation}\label{eqn:1}
error = \sum_{states} \Big(\frac{\left | GDP_{actual} - GDP_{predicted} \right |}{GDP_{actual}}*\frac{GDP_{state}}{GDP_{national}}*10\Big)
\end{equation}

The weight was multiplied by a constant (10) to ensure floating point errors did not occur and the numbers were of a comparable size. The results are shown in Table \ref{table: 1}.

\begin{table}[hbt!]
  \centering
  \caption{Reduction in weighted error of predicted GDP values after adding latitude and logitude  (2019)}
  \label{table: 1}
  \begin{tabular}{|c|c|c|}
  \hline
    Quarter & Only Nightlights & ReGNL (Nightlights + Lat, Long)\\
    \hline
    2019 Q1 & 7.3477 & 0.7697\\
    2019 Q2 & 5.7446 & 0.5034\\
    2019 Q3 & 5.8081 & 0.8681\\
    2019 Q4 & 5.9619 & 0.6852\\ \hline \hline
    Average & 6.2156 & 0.7066\\ \hline
\end{tabular}
\end{table}

As is observed in Table \ref{table: 1} we were able to obtain a significant increase in ReGNL's performance by adding the latitude and longitude to our feature set. This is inline with previous works that have highlighted the role geography plays in the economy of a region. The model is able to take advantage of this information and use it to help determine which states are economically better off.




\subsection{Model}
\label{4.3}
This section elaborates further on our model, ReGNL. Several Machine Learning algorithms such as Support Vector Regressors \cite{10.5555/2998981.2999003}, Linear (and Polynomial) Regression, XGBoost \cite{Chen_2016} as well as Neural Networks \cite{deeplearning-book, feedforward-nn} were tested and it was found that the Neural Networks provided better and more consistent results when compared to the others. The results obtained using the different models are presented in Table \ref{table: 5}. Given their superior performance, a Neural Network architecture was adopted for ReGNL.

\begin{table}[hbt!]
  \centering
  \caption{Comparisions of different algorithms in terms of their average weighted error of 2019-20 GDP predictions}
  \label{table: 5}
  \begin{tabular}{|l|c|}
   \hline
    Method & Error\\
    \hline
    SVR & 6.8221\\
    Linear Regression & 5.1546\\
    Neural Network & 0.7066\\
    XGBoost & 5.1436\\ \hline
\end{tabular}
\end{table}

\begin{figure}
  \centering
  \includegraphics[scale=0.85]{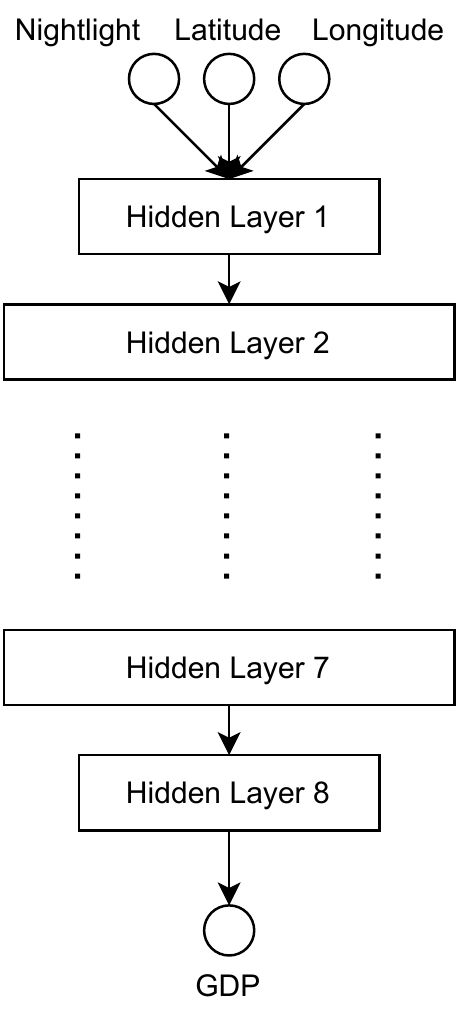}
  \caption{The Neural Network}
 \label{fig: 5}
\end{figure}

After running various tests with different architectures, based on train and test loss, the final model developed and trained was a feed forward neural network consisting of 8 hidden layers. The ReLU activation function was used after each layer (barring the last one) to introduce non-linearity. Regularization techniques such as dropout and weight decay were used to avoid over-fitting on the training data. The model was trained for $5x10^{6}$ epochs with a learning rate of $10^{-6}$. A schematic structure of the neural network is shown in Figure \ref{fig: 5}.

\begin{figure}
  \centering
  \includegraphics[width=\linewidth,height=\textheight,keepaspectratio]{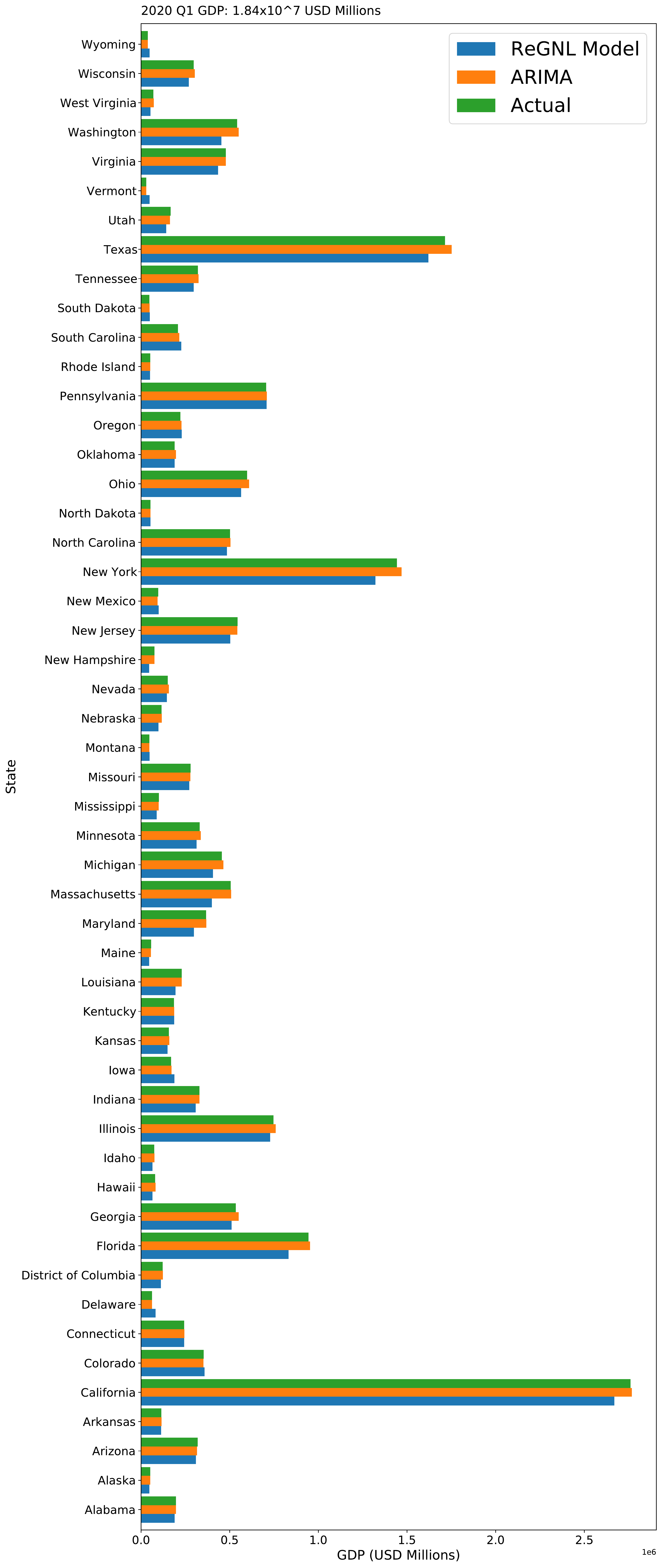}
  \caption{The predicted GDP (ARIMA and ReGNL model) values vs actual values for USA Q1 2020.}\label{fig: 3}
\end{figure}

Predictions for the year 2020 using all 3 features (nightlight, latitude and longitude) were made using ReGNL and the results of the same can be visualised in Figures \ref{fig: 3} and \ref{fig: 4} and their respective errors are shown in Table \ref{table: 2}. 

\begin{table}[hbt!]
  \centering
  \caption{Weighted error of predicted GDP values (2020)}
  \label{table: 2}
  \begin{tabular}{|c|c|}
   \hline
    Quarter & ReGNL (Nightlight, Lat, Long)\\
    \hline
    2020 Q1 & 0.7243\\
    2020 Q2 & 0.6879\\
    2020 Q3 & 0.6983\\
    2020 Q4 & 0.6478\\ \hline \hline
    Average & 0.6895\\ \hline
\end{tabular}
\end{table}


\begin{figure}
\label{fig: 4}
  \centering
  \includegraphics[width=\linewidth,height=\textheight,keepaspectratio]{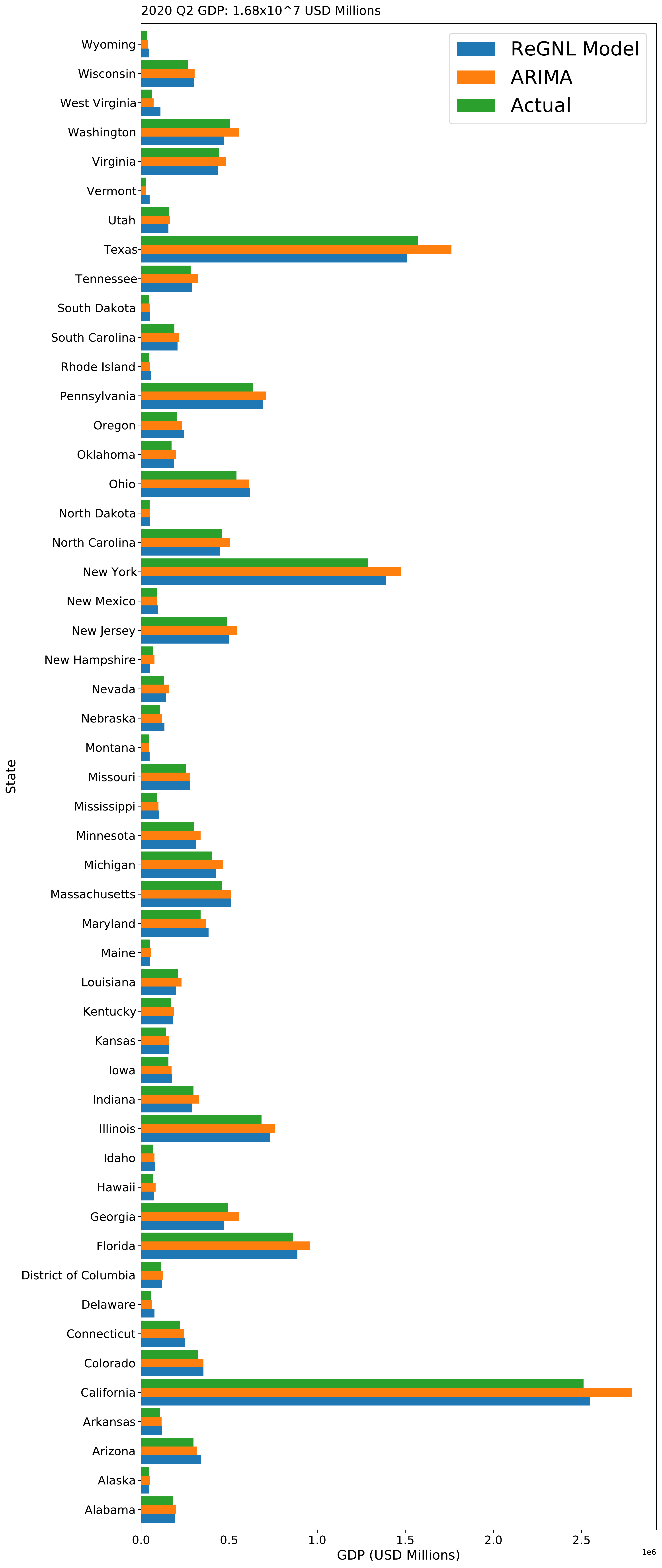}
  \caption{The predicted GDP (ARIMA and ReGNL model) values vs actual values for USA Q2 2020.}\label{fig: 4}
\end{figure}

\begin{figure*}[hbt!]
  \centering
  \includegraphics[width=\textwidth]{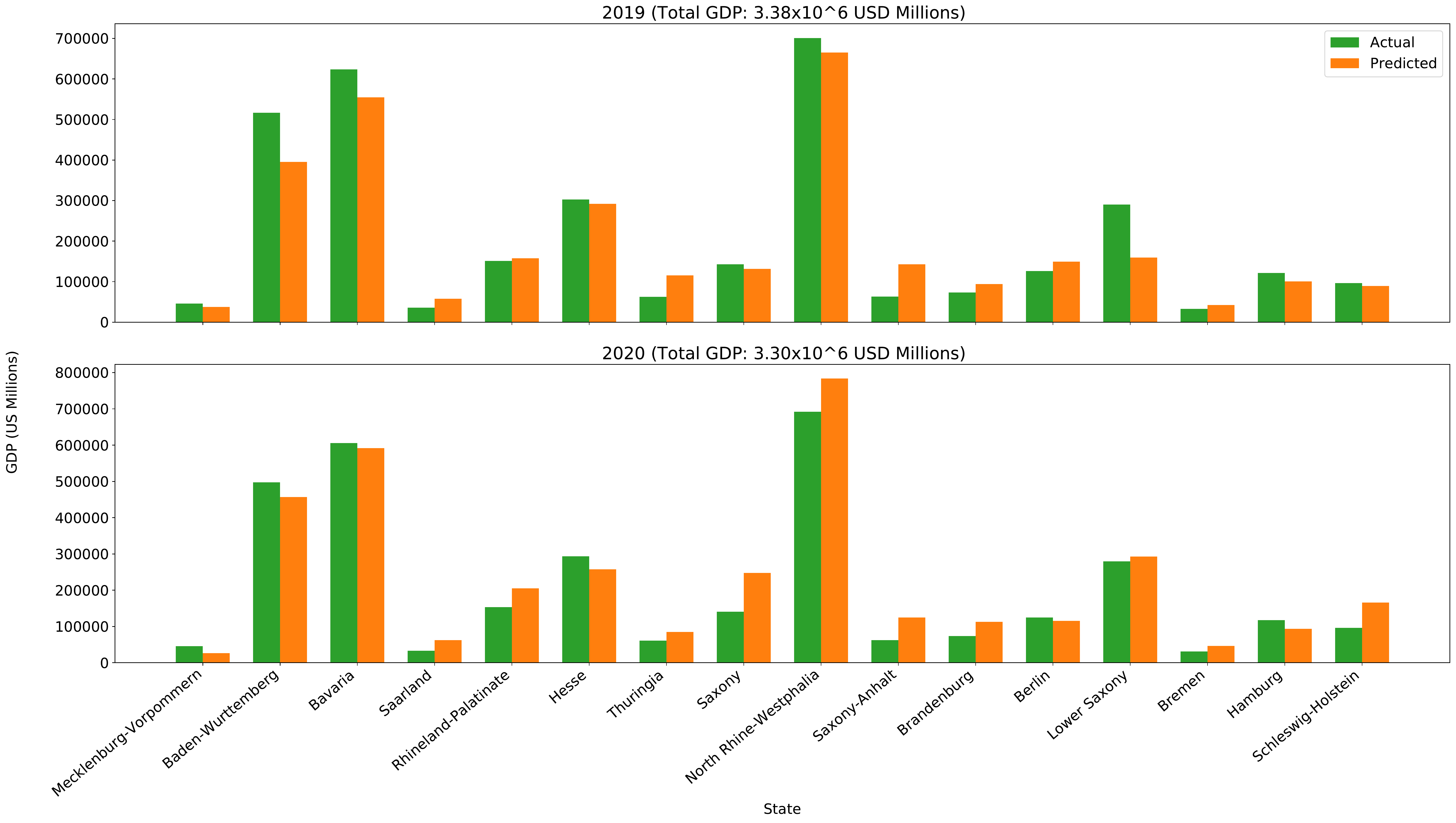}
  \caption{The predicted GDP values vs actual values for Germany (2019-20).}
  \label{fig: germany}
\end{figure*}

As can be seen from the table, the error obtained across the year was fairly consistent. The biggest takeaway is the low error from 2020 Q2. As shown in Figure \ref{fig: 3}, despite the GDP being hit drastically during this period, our model did not produce any anomalous result and continued to perform as hypothesized, in a similar fashion to as it was before the pandemic struck. These results provide evidence to support our claim that nightlight along with geographical data can act as a reasonable proxy to determine the GDP for the United States, even during disruptive events such as the COVID-19 pandemic. 

ARIMA is a technique used for time-series forecasting and has been used to forecast GDP values in previous works \cite{articlearima}. We employ ARIMA to predict the GDP in USA states for the years 2019 and 2020 as a baseline prediction estimate. We can also use this information to gain insight into whether the nightlight and geographical features of a region were able to serve as proxies to estimate the GDP more accurately. 
The results obtained through the ARIMA model are reported in Table \ref{table: 4}. A comparison of The actual and predicted GDP values using ARIMA and our model (ReGNL) is shown in Figure \ref{fig: 3} and Figure \ref{fig: 4}. It is clear that the ARIMA model overestimated the GDP values in 2020 Q2 whereas the estimations from the ReGNL model were closer to the ground truth. 

\begin{table}
  \centering
  \caption{Weighted error of predicted GDP values using ARIMA (2020) vs error of predicted GDP values using ReGNL}
  \label{table: 4}
  \begin{tabular}{|c|c|c|}
   \hline
    Quarter & ARIMA Error & ReGNL Error\\
    \hline
    2020 Q1 & 0.1235 & 0.7243\\
    2020 Q2 & 1.4976 & 0.6879\\ \hline
\end{tabular}
\end{table}


As can be observed, while the ARIMA model performs well in the first quarter, it is unable to take into account the impact of the pandemic and gives us predictions that are quite far off in the second quarter. Whereas ReGNL is able to perform much better during this time period and provides more consistent results even in the quarters affected by the disruptive event of the pandemic. ReGNL takes into account the nightlight along with geographical information and hence is able to make reasonably accurate predictions even during a disruptive event such as the COVID-19 pandemic, unlike ARIMA, which is a time-series based algorithm. 


\subsection{Performance in other countries}
\label{4.4}

We also created a dataset based on the annual GDP for other countries, namely, China, India, Spain and Germany. Out of them, the GDP data for 2020 to evaluate the model was available only for Germany. An identical approach was employed to predict the 2020 GDP for this country to understand if our method could be extended beyond the United States. However, only annual GDP data is available for Germany (while quarterly data is unavailable). In order to improve the number of data points, we combined the data from all the four countries in this experiment for the training phase. We were able to establish reasonable predictions for the pre-pandemic timeframe, without any disruptive event. The results for the same can be seen in Figure \ref{fig: germany} and Table \ref{table: 3}.  We note that the results obtained here are not as accurate as those obtained for the United States. Possible reasons for the same are lack of granularity in the available GDP data for Germany. Only annual GDP data was available for Germany, as opposed to the quarterly data available in the United States, which reduced the number of data points on which the model could be trained for Germany. Moreover, we pooled data from all these different countries i.e., China, India, Spain and Germany, to improve the number of data points, which might have affected the model. With sufficient and regular granular data, similar results as that of the USA can be obtained for these countries as well. 

\begin{table}[hbt!]
  \centering
  \caption{Weighted error of predicted GDP values for Germany}
  \label{table: 3}
  \begin{tabular}{|c|c|}
   \hline
    Year & ReGNL (Nightlight, Lat, Long)\\
    \hline
    2019 & 1.8649\\
    2020 & 1.9488\\ \hline
\end{tabular}
\end{table}


\section{Conclusions and Future Work}
\label{5}

Prediction of economic parameters, like GDP, becomes important in the case of disruptive events like the COVID-19 pandemic. However the economic parameters are not immediately available for the policymakers as the traditional methods to compute the parameters are time and resource intensive. In this paper we propose ReGNL, a neural networks based method, to be able to estimate the GDP of a geography using the nightlights data (obtained via remote sensing) in combination with geographical coordinates in the form of latitude and longitude. We are able to demonstrate that adding latitude and longitude as features for our predictive model improves the performance. Using USA as an example, we have also shown that the model is agnostic to the COVID-19 disruptive event. We have used ARIMA as a baseline for comparison and found that our model outperforms ARIMA for the timeframe when the disruptive event affected the economy. Additionally, we have presented a short extension of our work on Germany, in an attempt to build a model with a wider scope.

We believe one of the major bottlenecks in this space is the lack of available data. First, the VIIRS methodology for remotely sensing nightlight data has been deployed recently. Consequently, the VIIRS data is not available in abundance, putting a limit on the amount of training data we could use. The use of transfer learning could also be explored, potentially being able to fine tune for \emph{country A} when there is a pretained model for an economically similar \emph{country B}. Secondly, not all countries release their GDP statistics in the same format or with the same frequency, once again limiting the amount of data that could be used to construct a unified dataset. More research can also be done to understand the potential benefits of adding other remotely sensed indicators such as Carbon Monoxide emissions, NDVI (Normalized Difference Vegetation Index), etc. to the feature set to see if these can be used to predict economic parameters of different regions with improved accuracy. Finally, different model architectures can also be tried for the same task. The authors plan on releasing the dataset and source code publicly for the benefit of the scientific community. We want to make a case through this paper that countries should build robust data collection and publication frameworks which can help researchers and policy makers build models to understand the economic impact during disruptive events.



\bibliographystyle{ACM-Reference-Format}
\bibliography{sample-base}


\end{document}